\documentclass[conference]{IEEEtran}
\IEEEoverridecommandlockouts
\usepackage{cite}
\usepackage{amsmath,amssymb,amsfonts}
\usepackage{algorithmic}
\usepackage{graphicx}
\usepackage{textcomp}
\usepackage{xcolor}
\def\BibTeX{{\rm B\kern-.05em{\sc i\kern-.025em b}\kern-.08em
    T\kern-.1667em\lower.7ex\hbox{E}\kern-.125emX}}
\begin{document}

\title{Biometrics and Behavior Analysis for Detecting Distractions in e-Learning}

\author{\IEEEauthorblockN{
Álvaro Becerra\IEEEauthorrefmark{1},
Javier Irigoyen\IEEEauthorrefmark{1},
Roberto Daza\IEEEauthorrefmark{1},
Ruth Cobos\IEEEauthorrefmark{1},
Aythami Morales\IEEEauthorrefmark{1},
Julian Fierrez\IEEEauthorrefmark{1},
Mutlu Cukurova\IEEEauthorrefmark{2}}

\IEEEauthorblockA{\IEEEauthorrefmark{1}School of Engineering, Universidad Autonoma de Madrid, Spain\\
\{alvaro.becerra,
javier.irigoyen,
roberto.daza,
ruth.cobos,
aythami.morales,
julian.fierrez\}@uam.es}

\IEEEauthorblockA{\IEEEauthorrefmark{2}University College London, London, United Kingdom\\
\{m.cukurova\}@ucl.ac.uk}
}

\maketitle

\begin{abstract}
In this article, we explore computer vision approaches to detect abnormal head pose during e-learning sessions and we introduce a study on the effects of mobile phone usage during these sessions. We utilize behavioral data collected from 120 learners monitored while participating in a MOOC learning sessions. Our study focuses on the influence of phone-usage events on behavior and physiological responses, specifically attention, heart rate, and meditation, before, during, and after phone usage. Additionally, we propose an approach for estimating head pose events using images taken by the webcam during the MOOC learning sessions to detect phone-usage events. Our hypothesis suggests that head posture undergoes significant changes when learners interact with a mobile phone, contrasting with the typical behavior seen when learners face a computer during e-learning sessions. We propose an approach designed to detect deviations in head posture from the average observed during a learner's session, operating as a semi-supervised method. This system flags events indicating alterations in head posture for subsequent human review and selection of mobile phone usage occurrences with a sensitivity over $90\%$.
\end{abstract}


\section{Introduction}
In recent years, e-learning has grown exponentially, especially with online courses such as MOOCs (Massive Online Open Courses), which are often recognized by official institutions \cite{ma2019investigating}. However, e-learning environments face new challenges in maintaining high-quality learning standards due to their online nature. These challenges include the lack of direct contact between teachers and learners \cite{iraj2020understanding}, an increased likelihood of distractions during sessions (such as mobile phones, emails, conversations, etc.) \cite{winter2010effective, blasiman2018distracted}, and uncertainty regarding whether learners are paying attention to the content or experiencing difficulties \cite{daza2022alebk,daza2023matt}. 

This issue has prompted the development of Learning Analytics (LA) tools, offering valuable insights into online educational settings \cite{lang2017handbook, martinez2020achievements}, with numerous studies exploring learner behavior in MOOCs, analyzing patterns like demographics, motivation, and time spent on different types of content. They investigate how these patterns affect dropout rates and course completion \cite{rai2016influencing, kizilcec2017diverse}. Some studies focus on predicting learner performance using log data or other indicators \cite{moreno2018prediction}.

However, new e-learning platforms \cite{baro2018integration,hernandez2019edbb,daza2023edbb}, based on biometrics and behavioral analysis, have also emerged as a promising solution, benefiting from recent advances in machine learning and digital behavior understanding. The use of biosensors and Multimodal Learning Analytics (MMLA) tools aids in understanding learners' behavior and distractions \cite{giannakos2022multimodal}. Analyzing transitions between MOOC content and external sources helps predict learner success \cite{perez2019analyzing}. In \cite{rodriguez2024application}, mouse dynamics detect distractions (learners browsing the web), while \cite{betto2023distraction} uses facial features for the same purpose.

Nevertheless, in our days, navigating outside MOOC content isn't the only way to access external content. Several studies have focused on mobile phone usage during studying and analyzed them as a potential source of distractions \cite{david2015mobile}. In fact, researchers have attempted to find correlations between phone usage and the performance of learners. The majority of the results show a negative association after analyzing survey data from learners \cite{amez2020smartphone}.

In this article, we focus on detecting phone usage distractions. The main contributions of this work are: \textit{i}) An initial study on the effects of mobile phone usage during e-learning sessions, utilizing information obtained from the edBB platform \cite{hernandez2019edbb,daza2023edbb}, which was employed to monitor 120 learners during a MOOC Learning Session (LS). This study analyzes the impact of phone-usage events on behavior and physiological responses, specifically attention, heart rate, and meditation, which were studied before, during, and after phone usage.
\textit{ii}) A feasibility phone-usage events detection through video-based head pose estimation applied to e-Learning. 

The structure of this article is as follows: Section~\ref{s:related works} discusses related works. Section~\ref{s:LS} provides a detailed description of how biometrical data were obtained. In Section~\ref{s:methods}, we present the different computations, tests, and neural networks used for the analysis, and in Section~\ref{s:results}, we present the initial results. Finally, the article concludes with conclusions and future work (Section~\ref{s:conclusion}).

\section{Related Works}\label{s:related works}

With the increase in the number of users who own a phone \cite{smith2015us}, several research studies have been conducted to understand diversity in usage, addiction, and problems related to phone usage. Studies have shown that users interact with their phones an average of 10 to 200 times a day, with a mean interaction length ranging from 10 to 250 seconds \cite{falaki2010diversity}. The mean of general phone usage is approximately 162 minutes of which 32 minutes are spent using WhatsApp \cite{montag2015smartphone}.

Furthermore, phone addiction has become a wide research field \cite{atroszko2015study}. While several studies claim that there are correlations between addiction and stress or anxiety \cite{vahedi2018association}, other studies question whether it really is an addiction \cite{panova2018smartphone}.

In particular in the educational field, phone usage can be seen as a source of distraction that may negatively impact learner performance. In \cite{mendoza2018effect}, the researchers compared the performance and nomophobia levels of learners with different modes of phone usage (phone usage allowed, phone usage forbidden, and phone withdrawn). They found that learners with access to their phones showed worse performance in the later parts of the quiz, and there were associations between high levels of nomophobia and poorer performance.

While all these different studies rely on surveys, quizzes, and tests, information such as visual attention, cognitive load, or stress has demonstrated great value for the educational sector by helping understand how learners behave or where they are focused in order to predict their performance \cite{sharma2020eye}. Additionally, biometric information allows for the identification of possible suspicious and fraudulent events, as well as uncanny behaviors that instructors can observe to guarantee safe education or certification \cite{hernandez2019edbb,baro2018integration,morales2016kboc}.

Several studies \cite{giannakos2019multimodal,spikol2018supervised} have shown that better prediction results for learner success are obtained by using multiple data acquisition devices. An approach where multimodal data from multiple devices is monitored and analyzed in e-learning is the research presented in \cite{Azevedo2022} in the area of Intelligent Tutoring.

Recent advances in machine learning, especially in Convolutional Neural Networks (CNNs), have improved head pose estimation through video. State-of-the-art pose detectors, such as \cite{berral2021realheponet,ruiz2018fine}, have proven to be robust in challenging environments with changes in lighting, distance, etc. Considering these developments, and given that webcams are commonly integrated into personal computers, recent research has proposed methods for event detection in e-learning environments that leverage head pose detectors to accurately interpret and respond to learner actions and behaviors. Some examples are:  In this study \cite{daza2023matt}, the authors proposed estimating events of high or low attention through head posture estimation in conjunction with a Support Vector Machine (SVM). The authors of \cite{cote2016video} proposed a system to detect abnormal behavior events in learners using head pose estimations and a semantically meaningful two-state hidden Markov model.

\section{MultiModal Data Description}\label{s:LS}
To study how phone usage could affect learners while learning in a MOOC, we monitored 120 learners (59 females) from the School of Engineering at Universidad Autónoma de Madrid (UAM) who attended our MMLA laboratory and interacted with the same MOOC subunit during a 30-minute LS.  This study has been approved by the university's ethics committee, and all learner data are anonymized.

During the LS, the learners watch videos, read materials and answer assignments supported by the MOOC. In this initial research, we focused on 40 learners who were required to keep their phones visible during the LS, and they were instructed to use them upon receiving a notification. Each learner received at least two messages, as all learners were targeted with two messages from the researchers, and we labeled the period during which they were answering those messages as a phone-controlled event.

All MMLA data were collected using the edBB platform \cite{daza2023edbb}. It utilizes different sensors such as webcams, EEG band, smartwatches, eye trackers, etc. In our work, we utilized the following sensors:

\begin{itemize}
\item Video: Video data were captured from $3$ different positions: overhead, front, and side cameras, using $2$ simple webcams and $1$ Intel RealSense device that includes $1$ RGB and $2$ NIR cameras; the latter also calculates depth images using the NIR cameras. Additionally, screen monitoring video is recorded.
\item EEG data: Using a NeuroSky EEG band that obtains $5$ signals: $\delta$~($<4$Hz), $\theta$~($4$-$8$ Hz), $\alpha$~($8$-$13$ Hz), $\beta$~($13$-$30$ Hz), and $\gamma$~($>30$ Hz) and through the pre-processing of these EEG channels, attention (focus level) and meditation (mental calmness) are also obtained~\cite{daza2020mebal}. 
\item Heart rate: We use a Huawei Watch $2$ pulsometer feature~\cite{hernandez2020heart} to capture heart rate data.
\item Visual attention: A Tobii Pro Fusion which contains two eye tracking cameras that capture the following data: Gaze origin and point, pupil diameter, data quality, etc.; allowing us to obtain visual attention.
\end{itemize}

Furthermore, we store information related to the activities that the learner carries out during the LS, including timestamps for the start and end of each activity, along with a label for each activity (such as watching a video, reading, or using a mobile phone).

Thanks to the M2LADS system \cite{becerra2023m2lads}, all data captured from the LS were processed and synchronized to obtain information about the activity the learner was engaged in while each biometric value was being recorded.

\section{Methods}\label{s:methods}

\begin{figure*}[htbp]
  \centering
  \includegraphics[width=0.9\textwidth]{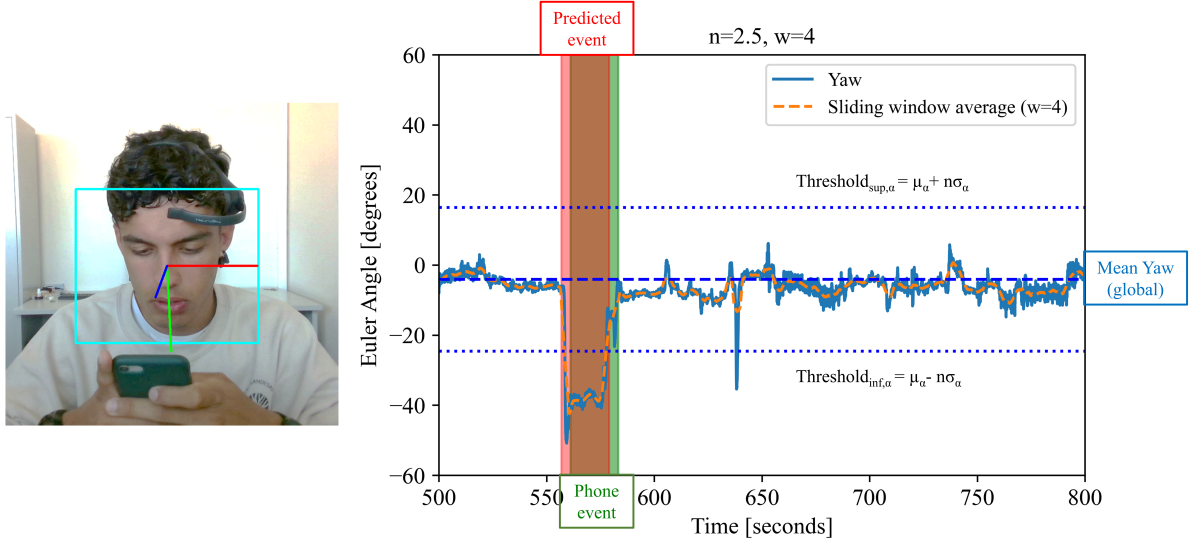}
  \caption{The head pose of a learner is captured at the instance of a phone event (left). The graph illustrates the yaw angle over time, captured during a LS (right). The solid blue line represents the raw data, the blue dashed line represents the global average and the orange dashed line indicates the local average, calculated using a sliding window of size \( w \). The dotted blue lines represent thresholds, adjustable via the parameter \( n \). When the local average crosses a threshold an event is predicted. Predicted events are denoted by a red shaded area, while actual events are denoted by a green shaded area.}

  \label{fig:fig1}
\end{figure*}
\subsection{Statistical Analysis}
Out of the 40 learners who used the phone during the LS, 33 (14 females) learners didn't present significant EEG data loss (more than 5 minutes of data loss), so 7 learners were discarded for this statistical analysis.

For each learner, we analyzed phone usage using NeuroSky meters to detect mental states, including attention (mental focus) and meditation (mental calmness), on a scale from 0 to 100, along with heart rate data from the smartwatch. Our focus was on phone-controlled events.

The average amount of time learners spent responding to each targeted message was $30.34$ seconds ($std=15.28$) with an average time for the first message of $25.42$ seconds ($std=10.97$) and $35.72$ seconds ($std=17.50$) for the second one. By gender, the average of time answering the messages is $29.95$ seconds ($std=13.63$) for men and $30.87$ seconds ($std=17.51$) for women.

We extracted the data for attention, meditation, and heart rate 15 seconds before the learner started using the phone, during phone-controlled usage, and 15 seconds after ending phone usage. We compared mean differences using t-Student test among the three periods for all learners and dividing them by gender.

\subsection{Head Pose Study}

We propose an approach to detect abnormal head pose events during e-learning sessions using webcam images. This approach was targeted at identifying events of mobile phone use. Our hypothesis suggests that head posture is significantly altered when interacting with a mobile device compared to the standard computer-based e-learning behavior. The approach aims to identify deviations in head posture from the session average, with a specific emphasis on adjusting parameters for the detection of mobile phone usage events. It operates as a semi-supervised method, flagging events for human intervention and selection of mobile phone usage events.

Two state-of-the-art modules, developed on CNNs, were employed to extract head poses. It includes the following:

\textbf{Face Detection module:} The facial detection module utilized the state-of-the-art MediaPipe's BlazeFace (full-range) model \cite{bazarevsky2019blazeface} to detect 2D facial images. This detector, an advanced version of BlazeFace, integrates an architecture inspired by MobileNet \cite{sandler2018mobilenetv2} for its feature extraction process, while its encoder is based on CenterNet \cite{duan2019centernet}. This combination enables a lightweight detection process, significantly enhancing its speed.

\textbf{Head Pose Module:} Head pose estimation was conducted using 2D facial images obtained from the facial detection module. Inspired by the WHENet model \cite{zhou2020whenet}, this module incorporates CNNs specifically designed for real-time head pose estimation. WHENet's architecture is based on an EfficientNet backbone \cite{tan2019efficientnet}, leveraging a strategic combination of classification and regression tasks to accurately predict the Euler angles. The head pose model was trained on the 300W-LP \cite{zhu2016face} and CMU Panoptic \cite{joo2015panoptic} datasets. This architecture calculates the vertical (pitch), horizontal (yaw), and longitudinal (roll) angles, enabling the deduction of 3D head orientation from 2D facial images (see Fig. \ref{fig:fig1}).

\textbf{Approach for Abnormal Head Pose Detection:} To detect abnormal head posture events from the LS, we adopted an offline processing strategy as follows:
\textit{i)} The Euler angles (pitch, roll, and yaw) were calculated for each frame. \textit{ii)} The mean ($\mu$) and standard deviation ($\sigma$) across the entire session were determined for each Euler angle. \textit{iii)} A sliding window methodology was adopted, through which the mean $(\overline{\theta})$ for each of the Euler angles was calculated within each temporal window $w$ (Equation \ref{eq:window}):



\begin{equation}
\overline{\theta}_i = \frac{1}{w} \sum_{k=s}^{s+w-1} \theta_{i,k}
\label{eq:window}
\end{equation}

\noindent where $\theta_{i,k}$ represents the angle $i$ (yaw, pitch, or roll) at frame $k$ and $s$ is the start frame of the window. (see Fig. \ref{fig:fig1})

\textit{iv)} A temporal window of $w$ frames is flagged as an event if the average value of any Euler angle within that window deviates beyond the threshold defined by the global mean $\mu$ and the global standard deviation $\sigma$ multiplied by a factor $n$ (Equation \ref{eq:Dazaequation}).

\begin{equation}
\begin{aligned}
&|\theta_i - \mu_i| > n\sigma_i \\
\end{aligned}
\label{eq:Dazaequation}
\end{equation}

\textit{v)} All overlapping events are merged into a single one to ensure that the events are disjoint.

Parameters $n$ and $w$ can be adjusted to enhance performance depending on the target event, enabling its adaptability to different scenarios.

\section{Results}\label{s:results}
\subsection{Statistical Analysis}

For all learners, higher attention levels were found after using the phone than before ($t(65) = 2.21, p = 0.031, d = 0.29$). Before using the phone, the average attention is $48.39\%$, and afterward, it increases to $51.61\%$ (Table \ref{tab1}). While in \cite{monsell2003task}, task switching seems to fall into efforts that could affect cognitive load, other studies \cite{salvucci2011toward} argue that outside a laboratory, these effects of multitasking are not really relevant. In our study, we didn't find evidence that multitasking (using the phone while learning) affects the 15 seconds after using it.

For all learners, lower heart rate levels were found before using the phone than during usage ($t(65) = 2.85, p = 0.006, d = 0.15$). Furthermore, this difference is particularly significant among men, who exhibited higher heart rate levels while using the phone ($83.24$ bpm) and afterward ($83.54$ bpm) compared to before usage ($81.26$ bpm) ($t(37) = 2.62, p = 0.013, d = 0.17$, and $t(37) = 2.5, p = 0.017, d = 0.20$). As learners had to answer text messages from the researchers, we hypothesize that this increase could be caused by the stress of responding to the messages and the possibility of receiving more, given that heart rate has been correlated with stress \cite{kim2018stress, choi2009using}. Additionally, other studies \cite{hooker2018just} show that female heart rates decrease after receiving a supportive message from their partners. However, as in our study, messages came from researchers, this effect was not found.

For male learners, higher meditation levels were found during phone usage than after phone usage ($t(37) = 2.54, p = 0.015, d = 0.46$). While using the phone, mental calmness in men reached $57.26\%$, while after it decreased to $52.82\%$.

\begin{table}[tbp]
\begin{center}
\caption{Summary of the Mean (M) and Standard Deviation (SD) for the different variables before, during, and after phone usage, grouped by all learners and by gender.}
\begin{tabular}{|c|c|c|c|}
\hline
\textbf{Biometrical} & \textbf{Before} & \textbf{During} & \textbf{After}\\
\textbf{variable} & M (SD) & M (SD) & M (SD)\\
\hline
Attention (all) & 48.39 (9.42) & 48.86 (10.30) & 51.61 (12.17)
\\
Attention (female) & 47.54 (9.94) & 46.07 (8.63) & 50.88 (12.76)
\\
Attention (male) & 49.01 (9.11) & 50.91 (11.03) & 52.14 (11.86)
\\
\hline
Heart Rate (all) & 83.01 (10.74) & 84.60 (10.43) & 84.71 (10.57)
\\
Heart Rate (female) & 85.48 (9.11) & 86.51 (8.90) & 86.35 (10.45)
\\
Heart Rate (male) & 81.26 (11.56) & 83.24 (11.31) & 83.54 (10.64)
\\
\hline 
Meditation (all) & 53.64 (9.38) & 56.11 (9.74) & 54.04 (8.92)
\\
Meditation (female) & 52.48 (8.14) & 54.55 (9.13) & 55.70 (9.05)
\\
Meditation (male) & 54.50 (10.22) & 57.26 (10.13) & 52.82 (8.74)
\\
\hline
\end{tabular}

\label{tab1}
\end{center}
\end{table}

\begin{figure*}[tbp]
  \centering
  \includegraphics[width=0.9\textwidth]{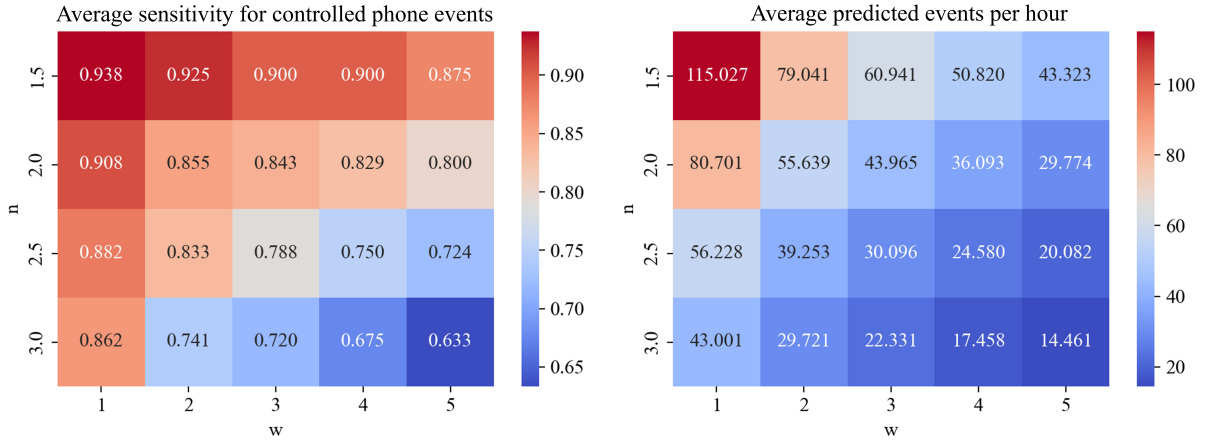}
  \caption{Average sensitivity obtained by the head pose based approach to detect phone usage across all users (left). Average number of events obtained by the head pose based the approach across all users (right). Axes $x$ and $y$ represent  parameters $w$ and $n$, respectively.}
  \label{fig:fig2}
\end{figure*}

\subsection{Head Pose Based Event Detection Approach for Phone Usage}

In this subsection, we present the findings from the application of our head pose based approach within the specific context of phone usage detection during a LS. The approach's performance was evaluated by testing it against labeled phone-controlled events, comparing them with the approach's predictions (see Fig. \ref{fig:fig1}). 

The sensitivity of the approach, reflecting its ability to correctly identify true positive events, was calculated across all 40 learners. These results are illustrated in Fig. \ref{fig:fig2} (left), evaluated across different parameter settings of $n$ and $w$. In this figure, it can be observed that higher sensitivity rates are observed with lower values of $n$ and $w$. For instance, a maximum sensitivity of 0.94 was achieved with $n = 1.5$ and $w = 1$. However, as $n$ increases, indicating a more restrictive criterion for event detection, sensitivity decreases. Similarly, increasing $w$ seems to reduce the approach's sensitivity.

The average number of predicted events was also determined for the corresponding settings of parameters $n$ and $w$. The outcomes are presented in Fig. \ref{fig:fig2} (right), illustrating the average number of events per hour. This metric provides further insight into the approach's performance, showing the effect of parameter adjustments on the frequency of event prediction throughout the LS. The results demonstrate an expected inverse relationship between the number of events detected and the parameters $n$ and $w$.

Identifying the optimal parameter settings can help maximize the performance for the detection of target events. Ideally, settings should ensure robust sensitivity while keeping event predictions at manageable levels. Combining insights from both tables, an example of such a balance could be achieved at $n = 2$ and $w = 5$ for the case of controlled phone events. Under this configuration, the approach would achieve a sensitivity of 0.80 with an average of 29 predicted events per hour. In practice, this translates to roughly 4 minutes within a 30-minute LS where 80\% of controlled phone events would be successfully captured, significantly optimizing the review process for supervisors overseeing the LS.

\section{Conclusion and Future Work}\label{s:conclusion}

In this article, we present: \textit{i}) An initial study on the effects of mobile phone usage during e-learning sessions, focusing on attention, heart rate, and meditation (mental calmness) before, during, and after phone usage.
\textit{ii}) A feasibility study on phone-usage events detection through video-based head pose estimation applied to e-learning. 

In our statistical analysis of attention, we noted a significant rise in attention levels among all learners following phone usage, compared to their attention levels before usage. Hence, in this preliminary investigation, we didn't find evidence suggesting that multitasking leads to decreased attention. Regarding to heart rate, we observed an increase in beats per minute during phone use compared to pre-usage levels. Notably, this disparity was more pronounced among men, who exhibited elevated heart rates both during and after phone usage compared to before. We theorize that this increase could be caused from the stress of responding to messages and the anticipation of receiving replies. Lastly, concerning meditation, we found that men reported higher levels of meditation during phone usage compared to afterward.

We also present an approach for detecting abnormal head pose events, with a feasibility study in an e-learning environment, focusing on phone usage events. Optimizing the approach's parameters for this task allows the identification of 80\% of target events while saving the need to review 86\% of the learning session. The results shown illustrate its potential as a tool for enhancing monitoring efficiency and time management. This approach could be also applied to other event types by readjusting its configuration.



As all of the significant differences in attention, heart rate, and meditation prove that phone usage causes changes in these biometrical variables. In future works, we will employ machine learning to detect phone events using these variables. Subsequently, we will integrate this predictor with the head pose-based event detection approach for phone usage to enhance its performance. Inclusion of RNNs\cite{shi2015convolutional} in the head pose module could refine the system given the sequential nature of the data. Furthermore, additional metrics such as body pose\cite{bazarevsky2020blazepose} and gaze direction\cite{chong2020detecting} could be integrated alongside head pose estimation to enrich the approach capability.

\section*{Acknowledgment}
Support: HumanCAIC (TED2021-131787B-I00 MICINN), SNOLA (RED2022-134284-T), e-Madrid-CM (S2018/TCS-4307), IndiGo! (PID2019-105951RB-I00), TEA360 (PID2023-150488OB-I00, SPID202300X150488IV0), BIO-PROCTORING (GNOSS, Agreement Ministerio de Defensa-UAM-FUAM dated 29-03-2022), and Cátedra ENIA UAM-VERIDAS en IA Responsable (NextGenerationEU PRTR TSI-100927-2023-2). Roberto Daza is supported by a FPI fellowship from MINECO/FEDER.

\bibliographystyle{IEEEtran}
\bibliography{bibliography}

\end{document}